\documentclass[letterpaper, 10 pt, conference]{ieeeconf}  
\IEEEoverridecommandlockouts
\usepackage{graphicx}
\usepackage{epstopdf}
\usepackage{amsmath}
\usepackage{amssymb}
\usepackage{subfigure}
\usepackage{multirow}
\usepackage{pbox}
\usepackage{cite}
\usepackage{algorithm}
\usepackage{algorithmic}
\usepackage{booktabs}
\usepackage{wrapfig}
\usepackage{bm}
\usepackage{url}
\usepackage{esvect}
\usepackage{xcolor}

\DeclareSymbolFont{extraup}{U}{zavm}{m}{n}
\DeclareMathSymbol{\varheart}{\mathalpha}{extraup}{86}
\DeclareMathSymbol{\vardiamond}{\mathalpha}{extraup}{87}

\newtheorem{problem}{Problem}

\DeclareMathOperator*{\argmin}{arg\,min}

\renewcommand{\baselinestretch}{0.95}

\begin{document}
%


\title{\LARGE \bf Interpretable Run-Time Prediction and Planning in\\ Co-Robotic Environments}
\author{Rahul Peddi  and Nicola Bezzo%
\thanks{Rahul Peddi and Nicola Bezzo are with the Department of Systems and Information Engineering and the Charles L. Brown Department of Electrical and Computer Engineering, University of Virginia, Charlottesville, VA 22904, USA. Email: {\tt \{rp3cy, nb6be\}@virginia.edu}}}


\maketitle



\begin{abstract}
Mobile robots are traditionally developed to be reactive and avoid collisions with surrounding humans, often moving in unnatural ways without following social protocols, forcing people to behave very differently from human-human interaction rules. Humans, on the other hand, are seamlessly able to understand why they may \textit{interfere} with surrounding humans and change their behavior based on their reasoning, resulting in smooth, intuitive avoiding behaviors. In this paper, we propose an approach for a mobile robot to avoid interfering with the desired paths of surrounding humans. We leverage a library of previously observed trajectories to design a decision-tree based interpretable monitor that: i) predicts whether the robot is interfering with surrounding humans, ii) explains what behaviors are causing either prediction, and iii) plans corrective behaviors if interference is predicted. We also propose a validation scheme to improve the predictive model at run-time. The proposed approach is validated with simulations and experiments involving an unmanned ground vehicle (UGV) performing go-to-goal operations in the presence of humans, demonstrating non-interfering behaviors and run-time learning.


\end{abstract}

%
\IEEEpeerreviewmaketitle

\section{Introduction}

Autonomous mobile robots are rapidly finding their way into our society in an increasing number of applications, including delivering packages in urban environments or deployed in warehouses assisting human workers. As these robots share space with humans, we must consider how they affect the human experience; that is, how surrounding humans behave in the presence of a moving robot. In many cases, these robots treat surrounding humans as stationary obstacles, often moving in unnatural ways, leaving the human solely responsible for learning to work around the robots.

\begin{figure}[t]
    \includegraphics[width=1\columnwidth,height=5cm]{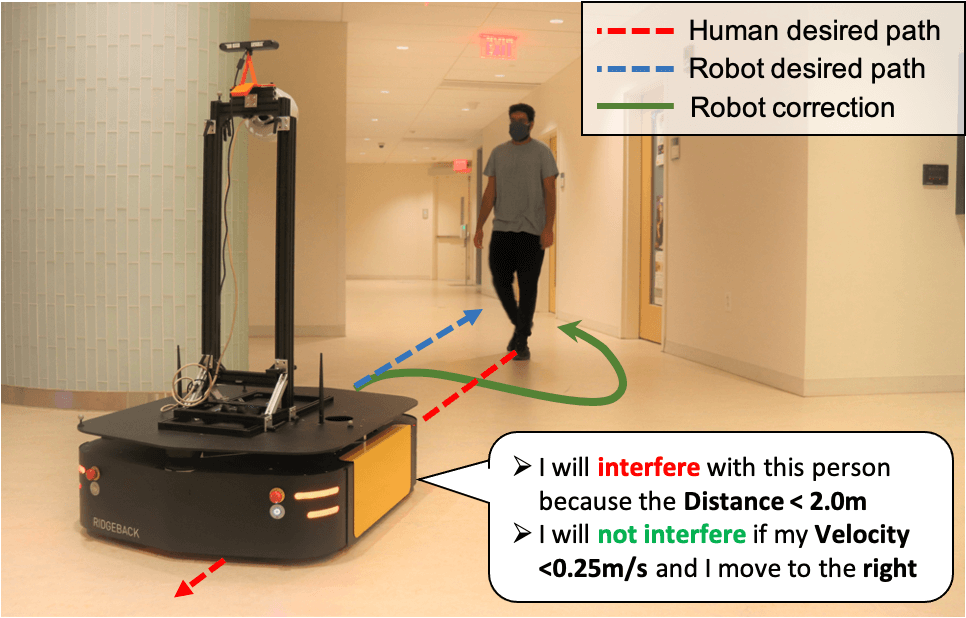}
    \vspace{-15pt}
    \caption {In our proposed approach, a robot predicts, explains and finds a corrective action to avoid interfering with an oncoming human.}
    \label{fig:intro}
    \vspace{-20pt}
\end{figure}

More recently, advanced machine learning techniques like Long Short Term Memory (LSTM) networks and Deep Reinforcement Learning (DRL) have been used to generate more natural robot behaviors around humans \cite{intentionlstm,cadrlmit,dnntraj}. While good robot behaviors can be produced, the approaches often contain black-box models \cite{blackbox}, and are unable to provide explanations or reasoning for decisions. In addition, these approaches typically are not adaptable at run-time and require dedicated training phases. 
Humans, on the other hand, accommodate others in very intuitive and easily interpretable ways. We are generally aware of our actions, and we can assess and explain if attributes of our behavior (e.g., how fast we are moving) will lead to some type of \textit{interference} with other people, causing them to change their path \cite{humanmotion}. We not only are able to explain whether we are interfering, but also intuitively use this explanation to change our motion-- without exactly predicting where others will go. If robots could reason about their behavior and plan corrective actions in a similar way, their motion would be easy to understand and interpret for surrounding humans. In Fig.~\ref{fig:intro}, we show a motivating example for this work in which a robot is able to predict, explain, and find a correction to avoid interfering with a person's intended path.

To achieve this behavior, we propose a novel method that leverages decision tree theory \cite{dtrees,carmelo} to predict, explain, and plan corrective actions at run-time in situations in which a robot will interfere with the motion of surrounding humans. Different from other learning-based approaches, besides the explainability aspect, another contribution of our work is that previously unobserved and misclassified data are considered at run-time through validation criteria to improve and refine predictions and corrective actions in future operations.

The rest of this paper is organized as follows: in Section~\ref{sec:relatedwork}, we discuss related literature. In Section~\ref{sec:probform}, we formally define the problem and in Section~\ref{sec:approach}, we describe our decision-tree based explainable monitoring and planning. In Sections~\ref{sec:sims} and~\ref{sec:exp}, we present simulation and experimental results and finally we draw conclusions and discuss future work in Section~\ref{sec:concs}.

\section{Related Work} \label{sec:relatedwork}

With recent advancements in mobile robots, there has been growing interest in enabling robots to navigate human environments in an intuitive and socially acceptable manner. Recent works in the field of machine learning have made substantial progress in enabling such behaviors, including~\cite{intentionlstm,rnntraj}, where the authors use recurrent neural networks (RNNs) or long short term memory networks (LSTMs) to predict motion of humans, which is used to generate safe robot motion. While these methods are effective for predicting human trajectories, they contain complex network architecture, and it is difficult to understand the mapping from input states to prediction. Authors in~\cite{irlmit} use deep reinforcement learning (DRL) to attain socially acceptable behaviors, and in~\cite{dnntraj}, the authors use DNNs to achieve similar results. While these approaches demonstrate good robot behaviors, none can provide explanations for why the resulting behavior was appropriate. In this paper, we complement the aforementioned works by interpreting and explaining predictions to generate non-interfering robot behaviors. 

In our more recent work~\cite{rahuliros}, we exploited Hidden Markov Model theory to obtain probabilistic predictions of temporal reachable states and developed a virtual physics-based planner, similar to the ``social-force'' developed in~\cite{socialforce} to operate the robot. In this paper, we bypass this requirement of explicit predictions and rely only on a more computationally efficient binary classification to achieve socially acceptable motion in co-robotic environments.

Towards explaining machine learning models, authors in~\cite{lime,lore} provide techniques to explain the predictions of a classifier. Instead of using global explanations, they find local reasoning as to why a data point was assigned to a certain class in simple classification learners, such as decision trees (DT). We take inspiration from these works and our previous work~\cite{carmelo}, where we have shown that DTs can be used to control a robot under bounded disturbance. We extend this work to handle a more dynamic environment considering multiple actors.

\section{Problem Formulation} \label{sec:probform}
Consider a mobile robot tasked to navigate an environment while avoiding other actors, in particular, humans. Without prior knowledge about the intended goal of the surrounding humans, this robot would not be able to predict their path. We note however that humans tend to move in certain way, typically in the direction of the desired goal. Let us define the path followed by a human as $\bm{q}^{*}_i$, with $ i=1,\ldots, N_h$ where $N_h$ is the number of humans in sensing range with the robot. With such premises we would like to design a framework for a robot to directly predict {\em interference} with all the humans in its sensing range and plan its motion accordingly to minimize the deviation of human paths due to its presence along the way.
Formally, the problem can be cast as:
\begin{problem}\label{prob:1}{\bf{Non-Interfering Motion Planning and Control.}} 
Design a policy to predict and explain future interfering interactions between a robot and surrounding humans and to plan corrective actions, $\bm{u}$, that do not cause human paths to deviate more than a distance $\delta$ from the intended trajectory:
\begin{equation}
||\bm{q}_h(t)-\bm{q}^{*}_h(t)|| \leq \delta, \forall h=1,\ldots, N_h(t), t\geq0
\end{equation}
where $\bm{q}_h(t)$ and $\bm{q}^{*}_h(t)$ are the observed path and intended path of the $i^{th}$ actor at time $t$ respectively.
\end{problem}

A correlated problem that we propose to investigate in this work is to improve robot behavior over time in response to previous experience. To this end, we create a strategy to validate and update predictions and planning at run-time when undesirable behaviors are observed or when run-time observations are unmodeled in the training data. 

\section{Approach} \label{sec:approach}

\begin{figure}[h]
    \includegraphics[width=1\columnwidth]{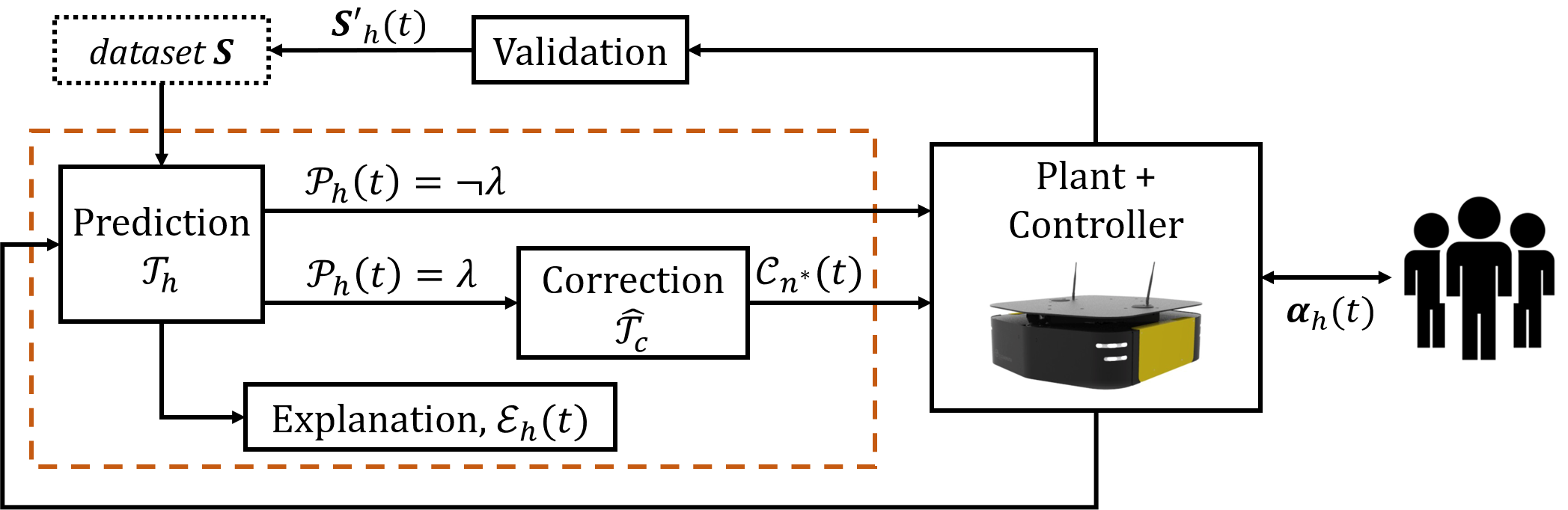}
    \caption{Block diagram of the presented approach.}
            \vspace{-5pt}
    \label{fig:block}
\end{figure}

Our proposed interpretable monitoring framework follows the architecture in Fig.~\ref{fig:block}.

\noindent At the core of our framework we leverage decision tree (DT) theory to predict interferences and correct robot behaviors. With observations of surrounding humans $\bm{\alpha}_h(t)$, a local DT, $\mathcal{T}_h$ is constructed with a dataset of human-robot trajectories to compute a prediction $\mathcal{P}_h(t)$ and explanation $\mathcal{E}_h(t)$ as to whether the robot will \textit{interfere} ($\lambda$) or \textit{not interfere} ($\neg\lambda$) with paths of surrounding humans. Then, when interference is predicted, a cascaded (secondary) tree $\hat{\mathcal{T}_c}$ is used to generate corrective behaviors that reduces robot interference with human paths. Finally, a validation scheme is proposed to update the dataset online, improving future DT operations. In the next sections, we describe in detail each component of our framework.

\subsection{Decision Tree Formulation and Training}
Decision trees (DTs) are a form of supervised learning that consist of white-box models which make predictions easy to interpret~\cite{dtrees}. In this work, DTs are constructed as binary classification models that are made up of a network of nodes; the outermost nodes, known as leaves, correspond to labels given in the training (decisions). Internal nodes define the split criteria for leaves based on the input variables (attributes). In this work, we grow DTs using the Gini Index as the splitting criterion (see~\cite{carmelo} for more details).

For training, we generate a dataset of trajectories in both simulation and in real experiments that consist of human motion from multiple initial to final positions with varying velocity in the presence of a moving robot. In the training, the robot does not react to the humans, so that the prediction model can learn when an interference occurs. In simulation, humans are controlled by a virtual physics-based method~\cite{rahuliros}, which triggers a reaction (interference) if a distance threshold, $\delta_{th}$, is violated. By training in this way, the desired effect is that the robot plans actions that keep a minimum distance of $\delta_{th}$ from all surrounding humans. 

Attributes should be meaningful to the application, and typically more attributes improve the precision of the prediction. However, too many attributes can lead to redundancy and poor classification~\cite{dtrees}. 
For the human-robot interaction case in this work, attributes are derived from the joint state between the robot and surrounding human, based on explicit sensor data available and implicit data we can compute (e.g., velocity from range sensor readings over time). 
Specifically, we define the attributes as
\begin{equation}
\bm{\alpha} = [~d_x~~d_y~~\theta~~d~~d'~~v_h~~v_r~~\ell~]
\end{equation}
where $d_x$, $d_y$, and $\theta$, are relative x-y positions and heading, respectively, $d$ and $d'$ are the Euclidean distance and distance derivative (i.e., the rate of change of the Euclidean distance) between human and robot, and $v_h$ and $v_r$ are the human and robot velocities, respectively. The robot's operating ``lane," $\ell$ is a discretization of the robot's $y$-position, assuming that the robot is typically moving forward, i.e., along the $x$-direction. 
 
\begin{figure}[h]
    \includegraphics[width=1\columnwidth]{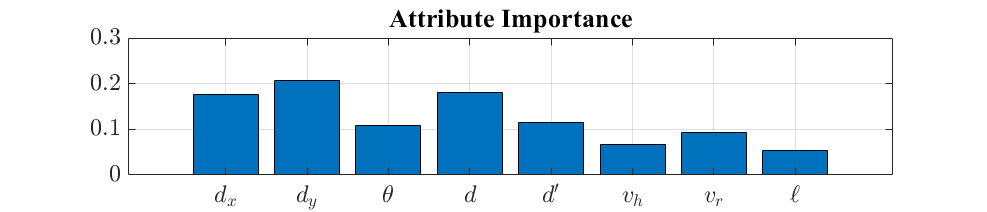}
    \caption{Attribute importance as a proportion of total attribute importance.}
    \label{fig:attimp}
\end{figure}

To validate the attributes, which were chosen via experimental sensitivity analysis, we analyzed the average predictor importance \cite{dtrees} of the attributes over 100 random local trees taken from subsets of the training data. A non-zero importance indicates that the attribute is valuable to decision tree predictions, and it is clear in Fig.~\ref{fig:attimp} that while some attributes may be more important than others, all attributes have an effect on the prediction. 

Through the training, we obtain a global dataset $\bm{\mathcal{S}} = \langle \bm{\alpha}_{\bm{s}},~\bm{\lambda_s} \rangle$ that includes both the attributes and corresponding classes of all training instances.

\subsection{Prediction and Explanation}  \label{sec:pred} \label{sec:explanation}
Let us now consider first the case of one human approaching the robot. To predict interference, we construct local DTs at run-time with the observed attributes related to a surrounding human, $\bm{\alpha}_h(t)$, since a global DT for the entire training set can often return inaccurate and imprecise predictions and explanations due to the presence of irrelevant data.
A local tree, $\mathcal{T}_h$, is trained by collecting a subset of points $\bm{\mathcal{S}_h(t)}\subset\bm{\mathcal{S}}$ from the global dataset that are within a neighborhood $\Delta$ of the attributes of $\bm{\alpha}_h(t)$:
\begin{equation}\label{eq:local_selection}
    \bm{\mathcal{S}_h(t)}\subset\bm{\mathcal{S}}\;| ~||\bm{\alpha}_h(t)-\bm{\alpha_s}|| \leq \Delta \;\; \forall \bm{\alpha_s} \in \bm{\mathcal{S}}
\end{equation}
The distance $\Delta$ is a measure of how close the local training data should be to the observed data. The exact value of $\Delta$ is selected based on the quality of the available training dataset. With a very rich dataset, a small $\Delta$ may result in very accurate predictions and explanations. For a sparse dataset, $\Delta$ should be large enough to ensure the decision tree has enough context to generate accurate predictions and explanations.
After constructing the local tree, the prediction $\mathcal{P}_h(t)\in[\lambda,\neg\lambda]$ is obtained by evaluating the run-time observation $\bm{\alpha}_h(t)$ in the tree: $\mathcal{P}_h(t) = \mathcal{T}_h(\bm{\alpha}_h(t))$.

Given a prediction, we compute an explanation $\mathcal{E}_h(t)$ by traversing the path through $\mathcal{T}_h$. A prediction directly corresponds to a leaf, $\mathcal{V}_p$, within the tree. If $\mathcal{V}_0$ is the root of $\mathcal{T}_h$, an explanation is computed by traversing a path, $\Gamma$ from $\mathcal{V}_0$ to $\mathcal{V}_p$, taking into account the split criterion, $c$, for the $N_i$ internal nodes along the path. The conjunction of split criterion along $\Gamma$ is the explanation of the prediction:

\begin{equation}\label{eq:explanation}
    \mathcal{E}_h(t) = \bigwedge_{k=1}^{N_i} {c}_k \quad \textrm{with}\quad \Gamma ~| ~\mathcal{P}_h(t)
\end{equation}

Traversing all other paths, $\Gamma_j\in\bm{q}$ with $j = 1,\ldots,N_{\bm{q}}$ that lead to the opposite decision, $\neg\mathcal{P}_h(t)$, in a similar way, provides a set of counterfactual rules, $\bm{\mathcal{C}}_h(t)$, to the previously obtained prediction:
\begin{equation}\label{eq:counterfactual}
\bm{\mathcal{C}}_h(t) = \bigvee_{j=1}^{N_{\bm{q}}} \bigwedge_{k=1}^{N_j} c_k \quad \textrm{with}\quad \Gamma_j ~| ~\neg \mathcal{P}_h(t)  
\end{equation}
where each path $\Gamma_j$ contains $N_j$ nodes to the leaf. These counterfactuals denote which attributes, if changed, would reverse the decision. 

\begin{figure}[h]
    \includegraphics[width=0.95\columnwidth]{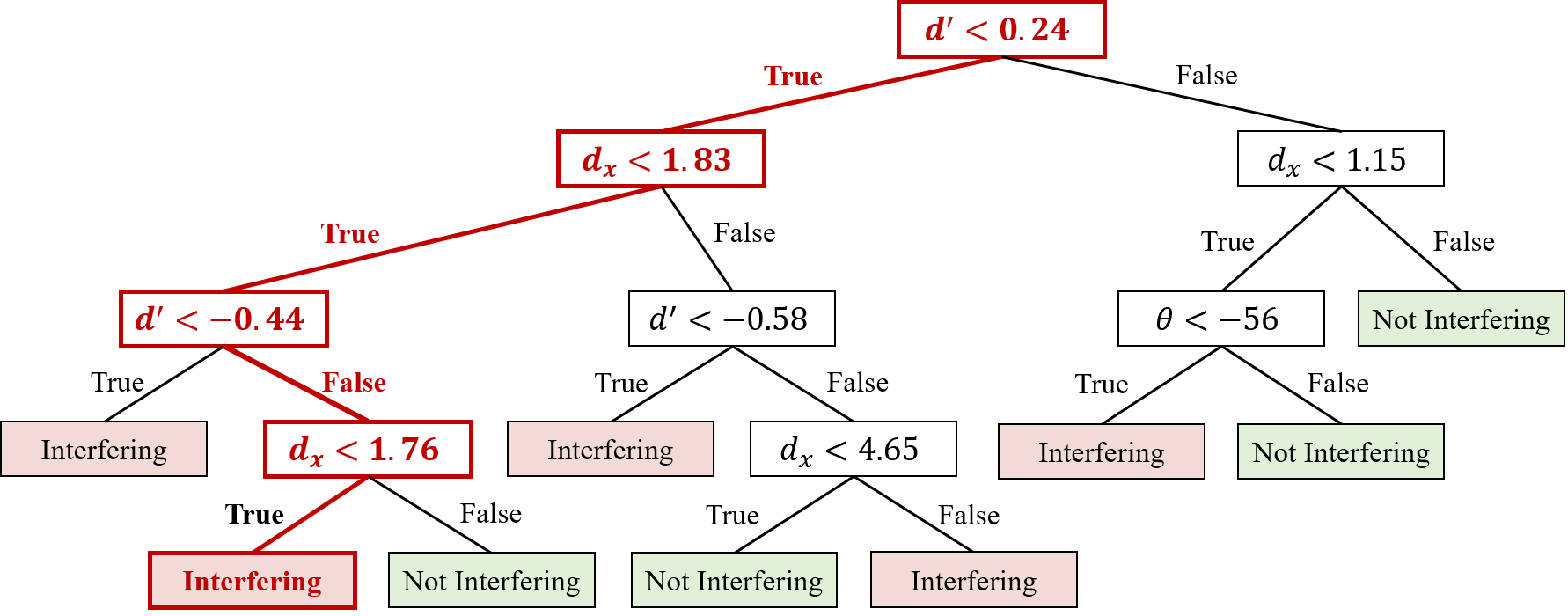}
    \caption{Example prediction DT. The internal nodes (white squares) of the tree are binary tests
on one of the attributes and the leaf nodes (colored squares) are the class decisions. The bold path shows the current decision.}
    \label{fig:extree}
\end{figure}

Shown in Fig.~\ref{fig:extree} is an example of a local DT used for prediction and explanation based on the following attributes, 
\begin{equation} \bm{\alpha}_h(t) =
\begin{bmatrix}
d_x \hspace{-2pt}&\hspace{-2pt} d_y \hspace{-2pt}&\hspace{-2pt} \theta \hspace{-2pt}&\hspace{-2pt} d \hspace{-2pt}&\hspace{-2pt} d' \hspace{-2pt}&\hspace{-2pt} v_h \hspace{-2pt}&\hspace{-2pt} v_r \hspace{-2pt}&\hspace{-2pt} \ell\\
1.43 \hspace{-2pt}&\hspace{-2pt} -4.71 \hspace{-2pt}&\hspace{-2pt} -81 \hspace{-2pt}&\hspace{-2pt} 4.93 \hspace{-2pt}&\hspace{-2pt} -0.43 \hspace{-2pt}&\hspace{-2pt} 1.0 \hspace{-2pt}&\hspace{-2pt} 0.6 \hspace{-2pt}&\hspace{-2pt} 0 \nonumber
\end{bmatrix}
\end{equation}
The output of the DT is $\mathcal{P}_h(t) = \lambda$, shown by the dark red path and leaf node. Through \eqref{eq:explanation} we compute an explanation:\\
\indent$\mathcal{E}_h(t)=$ \textit{\textbf{Interfering}} because: $\{d'<0.24, d_x<1.76 \}$\\
Through $\eqref{eq:counterfactual}$, we compute the following counterfactuals:\\
\indent$\bm{\mathcal{C}}_h(t)=$ \textit{\textbf{Not Interfering}} when:
\begin{itemize}
    \item[] $\{d'>0.24, d_x<1.15 \}~\lor$ 
    \item[] $\{\theta>-56, d'>0.24,d_x<1.15\}~\lor$
    \item[] $\{-0.45\leq d'<0.24, 1.76\leq d_x< 1.83\}~\lor$
    \item[] $\{-0.58\leq d'<0.24, d_x>1.83,d<4.65\}$
\end{itemize}

\begin{figure}[h]
    \includegraphics[width=1\columnwidth]{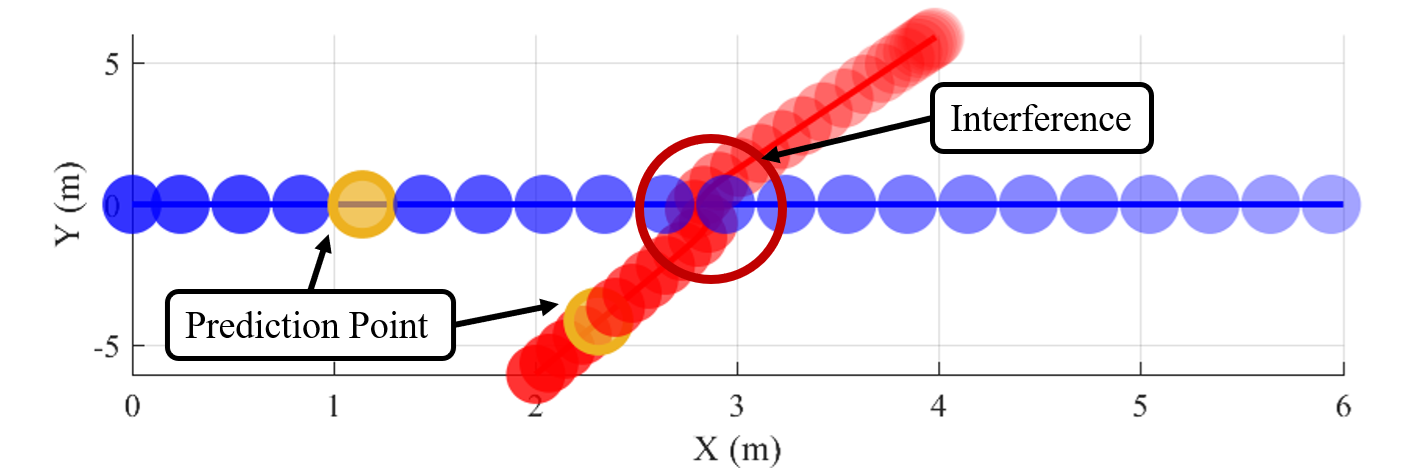}
    \caption{Human (red) and robot (blue) trajectories, showing the point (in yellow) at which a prediction is made.  The markers fade as time increases and the actors reach their goals.}
    \label{fig:predpoint}
    \vspace{-10pt}
\end{figure}

The point at which this prediction is made is highlighted in Fig.~\ref{fig:predpoint}, taken from our MATLAB simulations. 

As seen in the example, DTs used for prediction provide a logical and interpretable explanation, but the counterfactuals cannot be controlled by the robot's actions alone, as they pertain to human-dependent attributes.

\subsection{Corrective Counterfactual Analysis} \label{sec:correction}

In case of interference, the counterfactuals provide a set of configurations in which the robot would not have interfered with the path of the human. However, it is not practical to manipulate attributes like distance derivative $d'$ or relative heading $\theta$, since they depend on human motion.

The only controllable attributes in our case are the velocity and lane to track by the robot, $\bm{\alpha}_r = [v_r,~\ell]$. To generate actionable counterfactuals, we build a secondary tree, ${\mathcal{T}}_{c}$ in which we first fix the human dependent attributes $\bm{\alpha}_c=\bm{\alpha}_h\setminus\bm{\alpha}_r$ and search in the training set for similar attributes as done in \eqref{eq:local_selection} creating a new set $\bm{\mathcal{S}}_c \subset \bm{\mathcal{S}}$ (note that $\bm{\mathcal S}_h \subset \bm{\mathcal{S}}_c$). In this way, the new DT remains local in the human-related attributes but includes different lanes and velocity pairs, enabling the system to find suitable corrections. The new DT output in this way will be decisions and counterfactuals that only include $v_r$ and $\ell$.

\begin{figure}[t]
    \includegraphics[width=1\columnwidth]{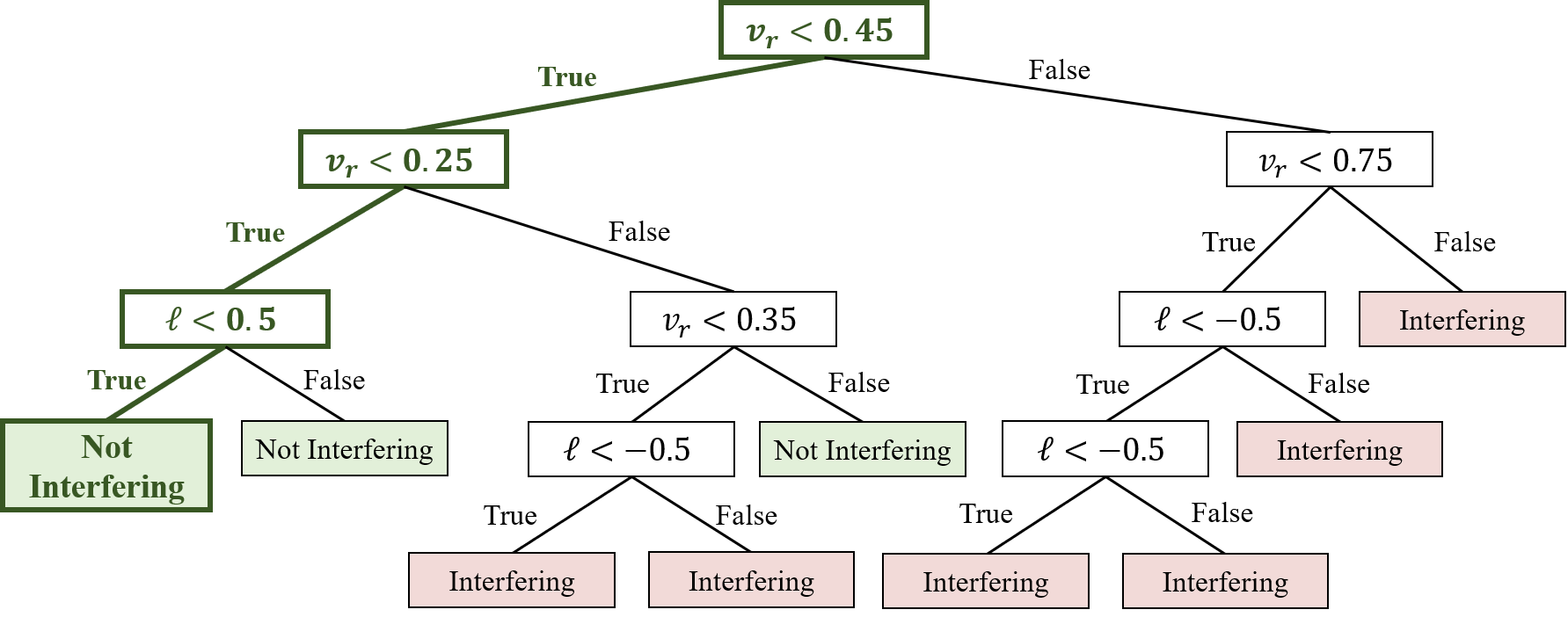}
    \caption{Example correction decision tree. Nodes and split criteria only pertain to $[v_r,~\ell]$. The bold path shows the optimal counterfactual $\mathcal{C}_{n^*}(t)$.}
    \label{fig:corrtree}
    \vspace{-15pt}
\end{figure}

In Fig.~\ref{fig:corrtree}, we show the correction tree associated with the example in Section~\ref{sec:explanation} and Fig.~\ref{fig:extree} in which the robot was initially running with $v_r=0.6$ and $\ell=0$. 

After running the procedure in this section, with \eqref{eq:counterfactual}, we obtain the following set of actionable counterfactuals:\\
\indent$\bm{\mathcal{C}}_h(t)=$ \textit{\textbf{Not Interfering}} when:
\begin{itemize}
    \item[] $\bm{\{v_r<0.25, \ell<0.5 \}}~\lor$
    \item[] $\{v_r<0.25, \ell\geq0.5\}~\lor$
    \item[] $\{0.35\leq v_r< 0.45\}$
\end{itemize}

To decide which counterfactual to select, we consider two measures that describe the quality of the nodes of the tree, \textit{node error}, $e_n$, and \textit{node risk}, $r_n$, with $n=1,\ldots,N_c(t)$, where $N_c(t)$ is the number of counterfactuals. Node error is the fraction of differently classified training points at a specific leaf. For a leaf that predicts $\neg\lambda$, the node error is:
\begin{equation}
e_n = 1 - p (\lambda)
\end{equation}
Node risk is a weighted measure of impurity (Gini Index in our work):
\begin{equation}
r_n = 1-\sum _{i=1}^{2}{\rho_{i}}^{2}
\end{equation}
where $\rho_i$ is the fraction of elements labeled with class $i = [\lambda,\neg\lambda]$. A lower node risk indicates that there will be less of chance of an incorrect decision. In our approach, we combine both measures by taking the product $e_n r_n$, since our goal is to identify the best node to use. The use of the product is viable here because both node error and node risk represent different probabilities that rely on information about the training points for the local tree, enabling the use of the general multiplication rule of probabilities for identifying the best node \cite{ranking}. Then, the optimal counterfactual is computed as follows:
\begin{equation}
    n^* = \argmin_{n}(e_n r_n)
\end{equation}
The selected counterfactual rule, $\mathcal{C}_{n^*}(t)$ consists of an optimal velocity and lane $\bm{\alpha^*} = [v_r^*,\ell^*]$. In the example shown in Fig \ref{fig:corrtree}, $\bm{\alpha^*} = \bm{\{v_r<0.25, \ell<0.5 \}}$.

\subsection{Corrective Planning and Control} \label{sec:planning}
Once a counterfactual rule is selected, the robot moves to implement the correction, which is represented as a ``ghost" moving target, and the robot switches into a pure-pursuit based mode of operation \cite{purepursuit} until it reaches the ghost vehicle. This is necessary because the corrective action represents what the robot should have been doing at the instance $t$ at which the correction was found, meaning that the robot would only satisfy non-interfering conditions if $v_r(t) = v_r^*$ and $\ell(t) = \ell^*$.

During the pure-pursuit corrective operation, the robot uses the ghost vehicle's state to make predictions until the tracking error between robot and ghost $e(t) = \bm{p}_g(t)-\bm{p}_r(t) \approx 0$, after which it reverts to performing predictions based on the actual state of the robot. This is needed because as the robot performs corrective behaviors, we observe transition states that are unmodeled in the training data, since the robot does not react to the humans in the training. 

\subsection{Multiple Decision Trees}

In this section, we discuss how our approach extends to scenarios with multiple actors. Predictions and explanations can be computed as discussed previously, as the system needs to understand if it's interfering with each actor individually. Finding corrections, however, is more challenging, as the corrective action must not only remove interference with one actor, but also should not cause interference with others. 

\begin{figure}[h]
    \includegraphics[width=0.99\columnwidth]{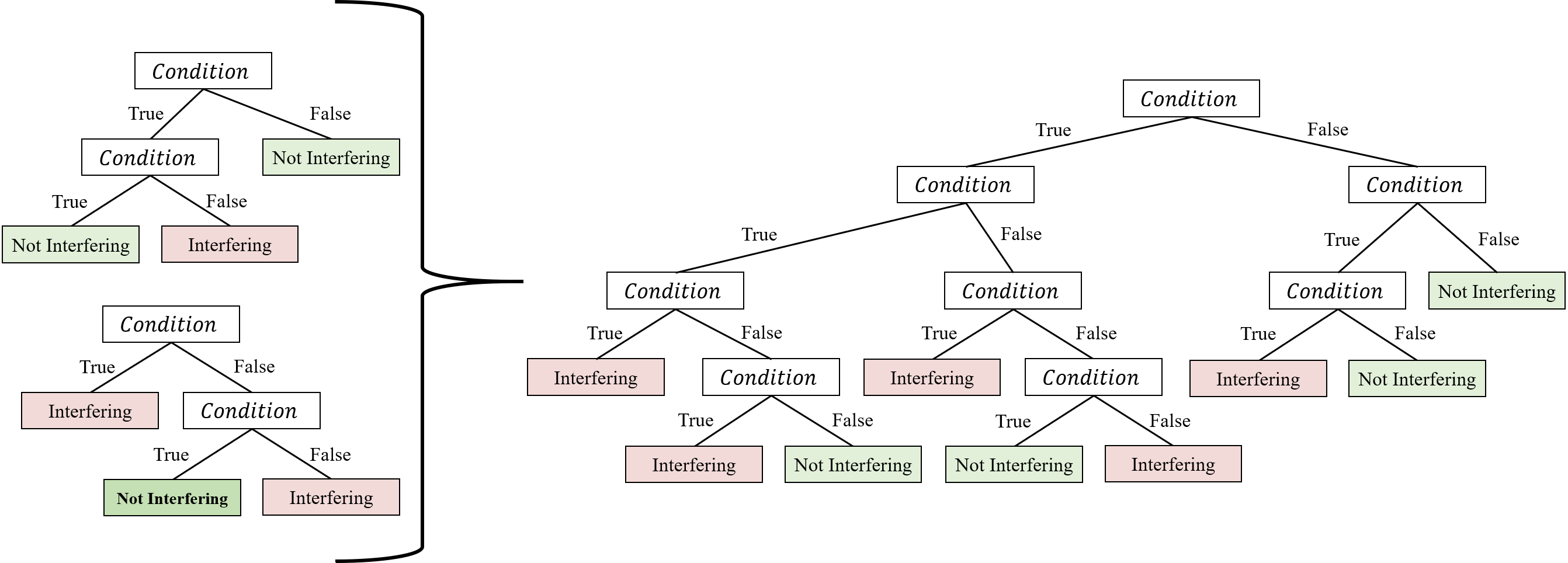}
    \caption{General example of a Decision Tree Ensemble Model. Two weak learners come together to form a stronger learner.}
    \label{fig:dtem}
\end{figure}

To consider different counterfactual rules of multiple DTs at once, we use {\em decision tree ensemble models} (DTEM). The main principle behind DTEM (Fig.~\ref{fig:dtem}) is that a group of weak learners come together to form a strong learner.
Some popular methods for generating DTEMs include bagging, boosting, or using random forests, but these consist of random sampling methods~\cite{ensemble}, which are not viable for our case, since we need to capture all relevant data.

We instead take principles of majority voting DTEMs \cite{majorityensemble}, which typically compare predictions from each DT, and select the majority output. We extend this concept by considering a single combined corrective tree, $\hat{\mathcal{T}}_c$, that incorporates the local training data of all individual trees, rather than just the prediction.

Before combining, the local data for each person are normalized such that $|\bm{\mathcal{S}}_i(t)| \approx |\bm{\mathcal{S}}_j(t)| ~\text{with}~ i,j = 1,\ldots,N_h$, where $|\cdot|$ gives the size of the enclosed dataset. In this way, corrections will not be incorrectly biased towards those with more local data. We combine to obtain the training dataset: $\hat{\bm{\mathcal{S}}}(t) = [\bm{\mathcal{S}}_1(t),\ldots,\bm{\mathcal{S}}_{N_h}(t)]$,
with which $\hat{\mathcal{T}}_c$ is constructed and counterfactuals are analyzed with the procedure discussed in Sec.~\ref{sec:correction}, to obtain the optimal target, $\bm{\alpha^*}$, and the robot is controlled as described in Sec.~\ref{sec:planning}.

Building $\hat{\mathcal{T}}_c$ in this way, however, only enables the system to find the best action if one exists within the data, meaning that there can be cases where all considered corrective actions are interfering. This can happen if: 1) the considered corrective data are sparse, meaning that the training set is not rich enough or 2) all possible actions are not feasible, for example, if the robot is surrounded by a crowd. For the former case,
we propose a randomized approach to choose a velocity-lane pair that is not included in the local data: $\bm{\alpha}_s\setminus\hat{\bm{\alpha}(t)}\in\hat{\bm{\mathcal{S}}}(t)$.
By doing so, the vehicle explores new options until it finds a solution. If no solution is found, the robot switches into a fail-safe mode of operation, which consists of a very low velocity and a reactive obstacle avoidance behavior for safety.

\subsection{Online Validation and Updating}  \label{sec:val}
Due to the dynamic and dense nature of the environment, new and unmodeled human behaviors can be observed and corrective actions may not always eliminate all interference. Since our DTs are constructed at run-time, it is possible to introduce new data to the training set $\bm{\mathcal{S}}$ as observations are made. 
To avoid an exploding dataset, we introduce two test cases for adding new data: 1) decision validation, and 2) checking for unbounded observations.

\textbf{Case 1}
Decision validation is necessary when corrective actions from $\hat{\mathcal{T}}_c$ still result in an interference. This can occur when observations contain attribute values that are close to splitting conditions in the DTs leading to misclassification, or when a correction cannot be found within the DTEM. If the robot observes that the distance threshold $\delta_{th}$ is violated at run-time, the recorded attributes are included in $\bm{\mathcal{S}}$.

\textbf{Case 2}
If an observation is outside the bounds of the training data, reliable predictions or corrections cannot be expected. It has been shown for learning components that testing data within the vertices of the smallest convex set around training data produces the most accurate predictions~\cite{inhull}. In this work, convex hulls are generated around local training data using the Quickhull algorithm~\cite{quickhull} to form a boundary denoted $Conv(\bm{\hat{\mathcal{S}}})$. Then, we check if observations are within the outermost points of each dimension (attribute) of the convex hull using linear inequalities~\cite{linearconst}:
\begin{equation}
    \min(Conv(\bm{\hat\mathcal{S}})) \leq \bm{\alpha}_h(t) \leq \max(Conv(\bm{\hat\mathcal{S}})) 
\end{equation}
If any part of $\bm{\alpha}_h(t)$ is outside the convex hull, the data are labeled and included in $\bm{\mathcal{S}}$. The dataset updates at run-time through the presented test cases, resulting in more refined local trees, and therefore better decision making in the future.

\section{Simulations} \label{sec:sims}

We performed a series of simulations in MATLAB to test the effectiveness of our approach. Training included velocities $\bm{v_r} = [0.2,0.4,0.6,0.8,1.0]$m/s, and lanes $\bm{\ell} = [-2,-1, 0, 1,2]$m simulating a classical non-holonomic UGV. The robot considers humans within a range of $5$m and has a nominal velocity of $0.6$m/s, and corrections are limited to any discrete velocity or lane seen in the training, that is from $\bm{v_r}$ and $\bm{\ell}$, respectively. 

\begin{figure}[ht!]
    \vspace{-10pt}
	\centering
	\subfigure[Paths of human(red) and robot(blue). \label{fig:ex1}]{\includegraphics[width=\columnwidth]{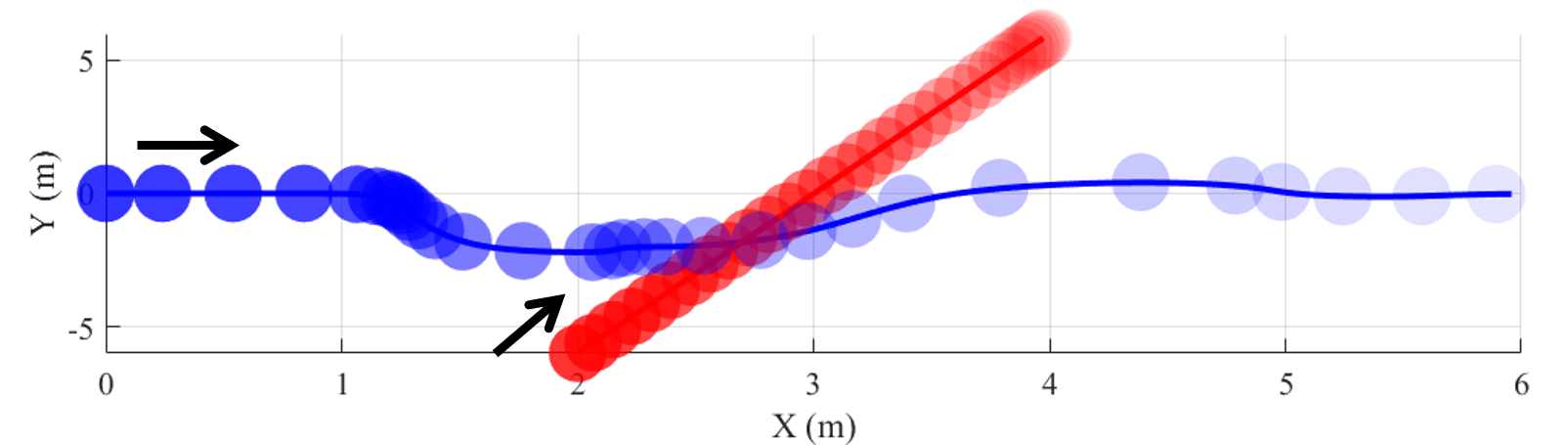}}
	\subfigure[Decision tree operations. \label{fig:ex0}]{\includegraphics[width=0.48\columnwidth]{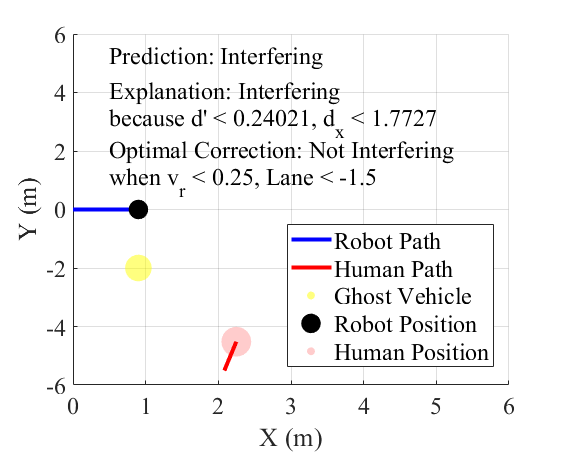}}
	\subfigure[Distance to human. \label{fig:ex3}]{\includegraphics[width=0.48\columnwidth]{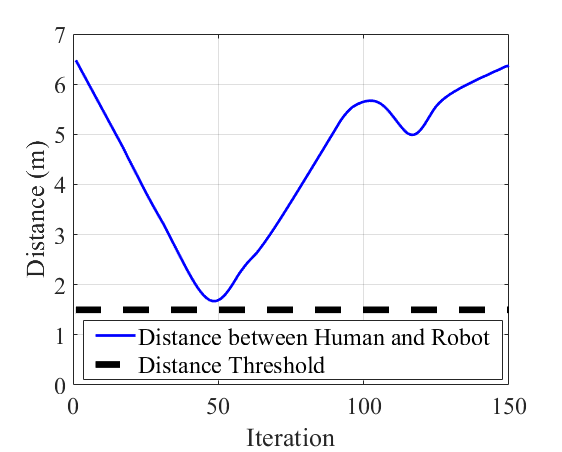}}
	\caption{Baseline simulation.}
	\vspace{-5pt}
	\label{fig:sims}

\end{figure}

In the baseline simulation shown in Fig. \ref{fig:sims}, the robot (blue) is tasked to move from $(0,0)$m to $(6,0)$m while predicting, explaining, and correcting to avoid interfering with a person (red) moving along a trajectory previously unknown to the robot (i.e., different from the training set).
The paths are shown in Fig.~\ref{fig:ex1}, and prediction, explanation, optimal correction, and ghost vehicle are shown in Fig.~\ref{fig:ex0}. The robot predicts $\mathcal{P}_h(t) = \lambda$ and determines that it must apply the correction: $\bm{\alpha}^* = [v_r < 0.25, \ell < -1.5]$. The ghost vehicle (yellow marker) immediately applies these corrections.
In Fig.~\ref{fig:ex3}, the distance between robot and human is shown to verify that the distance threshold $\delta_{th}=1.5$m is not violated.

\begin{figure}[h]
	\centering
	\subfigure[Paths of humans (red, magenta) and robot (blue). \label{fig:s0}]{\includegraphics[width=0.5\textwidth]{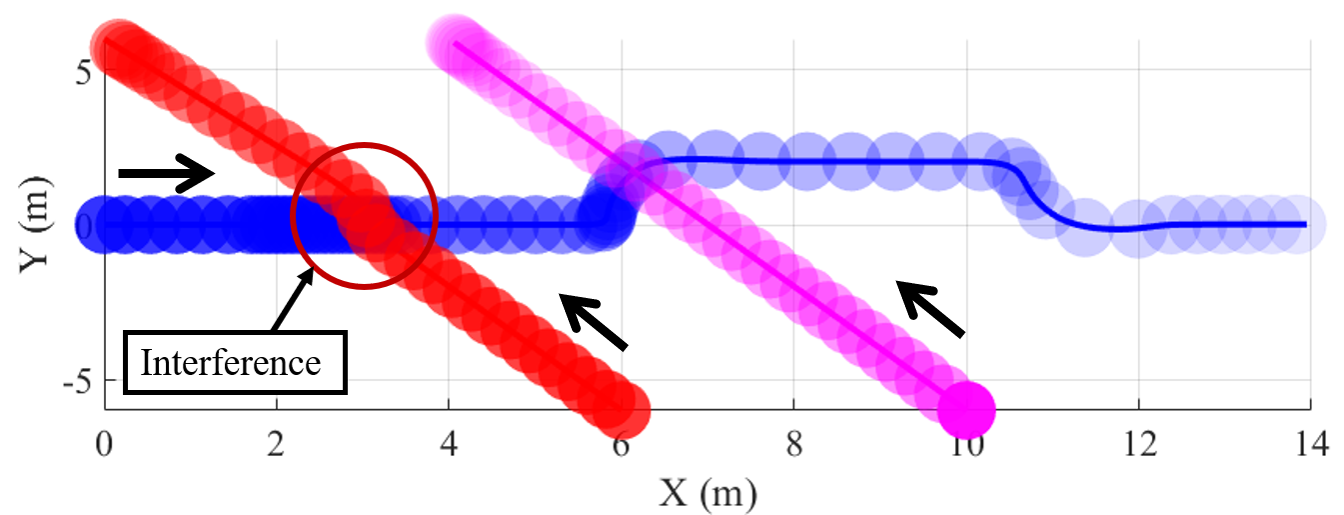}}
	\subfigure[Incorrect DT operations. 
	\label{fig:s1}]{\includegraphics[width=0.23\textwidth]{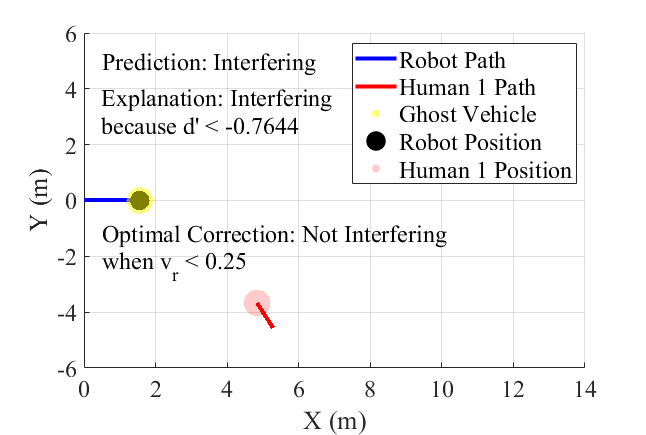}}
	\subfigure[Corrected DT operations. 
	\label{fig:s2}]{\includegraphics[width=0.23\textwidth]{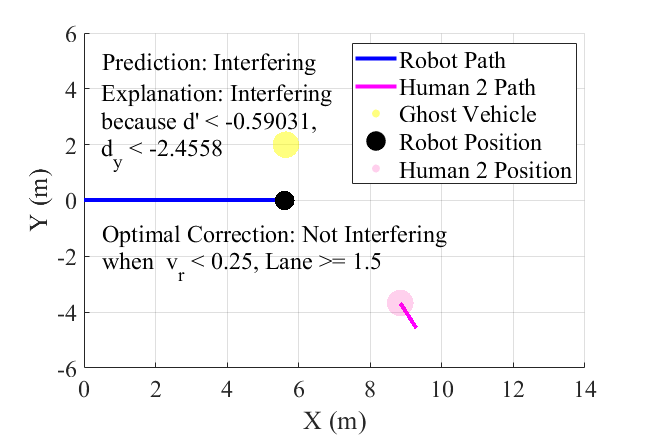}}
	\subfigure[Partial convex hull prior to updates. \label{fig:s3}]{\includegraphics[width=0.23\textwidth]{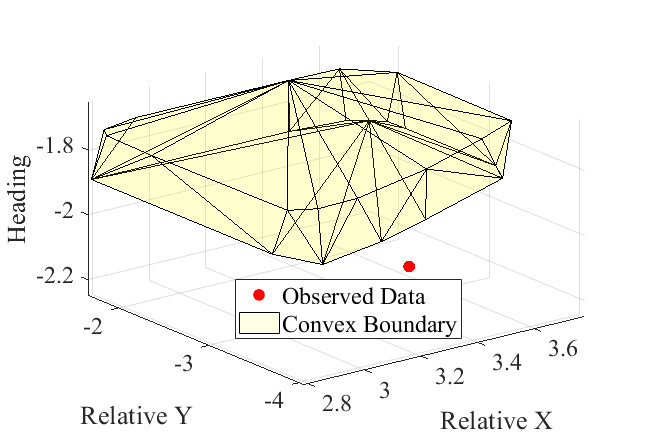}}
	\subfigure[Partial convex hull after updates. \label{fig:s4}]{\includegraphics[width=0.23\textwidth]{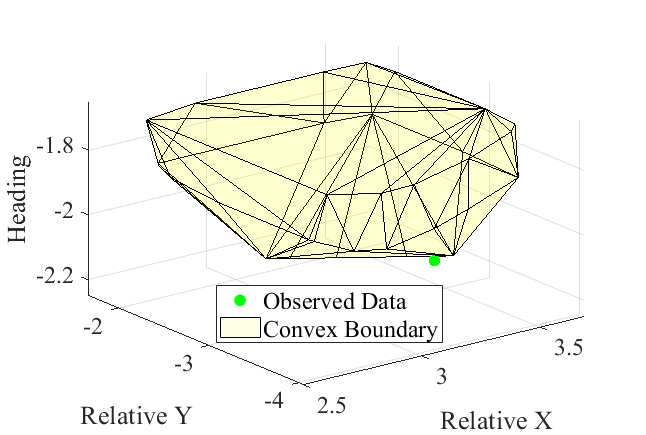}}
	\caption{Simulation of run-time validation and updating.}
	\label{fig:sims2}
\end{figure}

In Fig.~\ref{fig:sims2}, we show the effects of online validation and learning by performing a simulation with DTs trained on an incomplete training dataset. The robot's goal is $(14,0)$m and the humans in this simulation take identical paths in succession to test whether the robot has improved its behavior.
The robot makes an incorrect decision at first, applying the explanation and correction shown in Fig.~\ref{fig:s1}, only slowing down to $v_r<0.25$m/s. Both test cases (Sec.~\ref{sec:val}) are violated, shown by the deviation in the red path of Fig.~\ref{fig:s0} and the observation (red point) outside the partial convex hull in Fig.~\ref{fig:s3}. Note that partial $3$-dimensional convex hulls are shown for visualization purposes, due to the high dimensionality of our attributes, $\bm{\alpha}\in\mathbb{R}^8$. The magenta path in Fig.~\ref{fig:s0} shows no interference, as a different correction was selected (Fig.~\ref{fig:s2}), since the observations are now included in the local data (Fig.~\ref{fig:s4}).

\begin{figure}[h]
    \includegraphics[width=0.99\columnwidth]{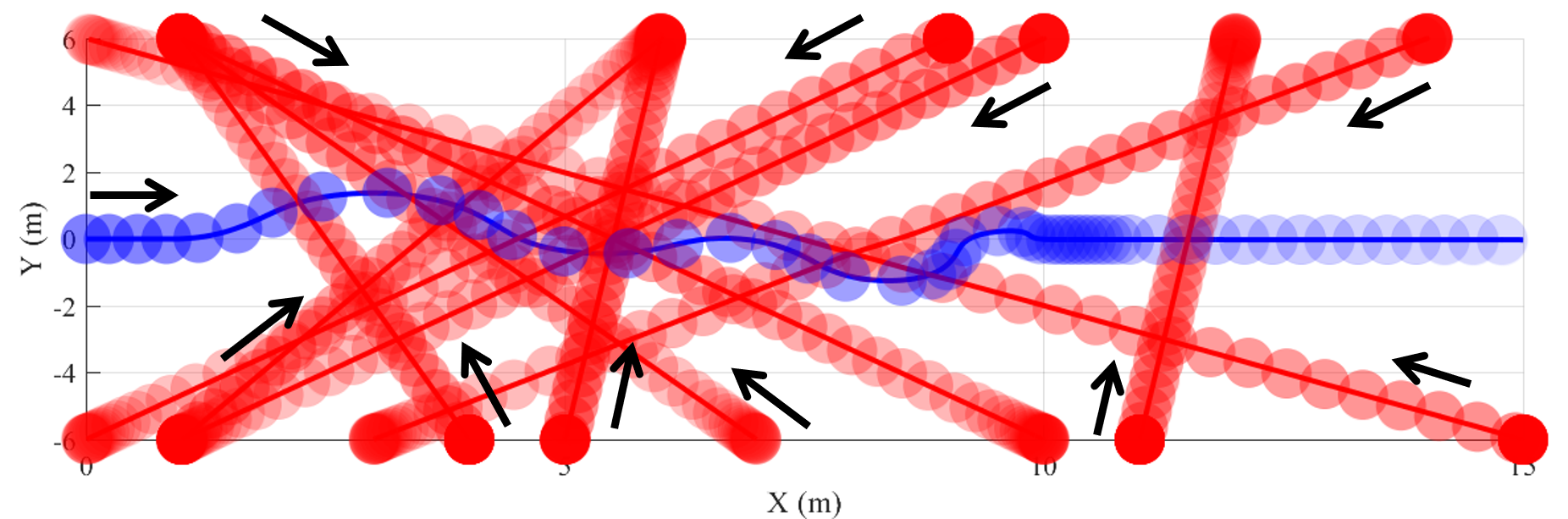}
    \caption{Human (red) and robot (blue) paths in a multi-actor simulation}
    \label{fig:bigsim}
    \vspace{-10pt}
\end{figure}

We also extensively test our approach in handling multiple people at a time, shown in Fig.~\ref{fig:bigsim}.
The robot navigates through 10 people, changing its lane ($\ell$) and velocity ($v_r$) to avoid interfering. This test was run for $50$ trials with random human trajectories. All decision tree operations in these simulations took between $30-90$ms. Our approach successfully eliminated interferences $72\%$ of the time, with an average minimum distance of $\delta_{\min}=1.58$m. Where interference occurred, we observed that the robot was in dense crowds, negotiating with on average $N_h(t)\geq 7$.

\section{Experiments} \label{sec:exp} 
The proposed approach was also validated experimentally on a Clearpath Robotics Ridgeback Omnidirectional Platform (see Fig.~\ref{fig:intro}) in indoor environments. In the first experiments, a VICON motion capture (MOCAP) system was used to obtain robot and human states. In the second experiment, the robot uses an on-board ASUS Xtion RGB-D camera with the SPENCER people tracking package \cite{spencer}. Below we present a few cases that capture the essence of the proposed framework. Videos of the presented experiments can be found in the submitted supplemental material.

\subsubsection{MOCAP Experiments}

\begin{figure}[h]
    \centering
	\subfigure[Snapshots of experiment.
	\label{fig:exp1snap}]{\includegraphics[width=0.5\columnwidth]{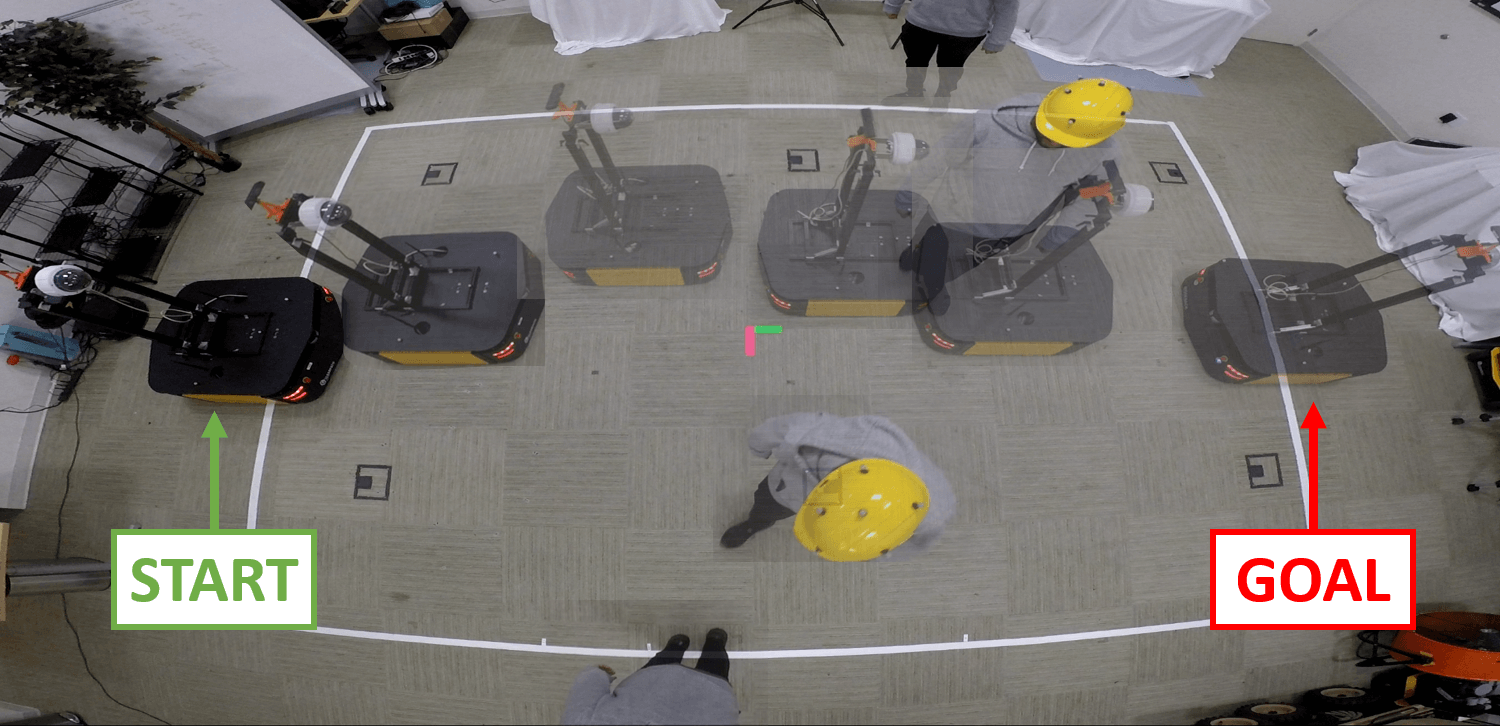}}
	\subfigure[Distance between actors.
	\label{fig:exp1dist}]{\includegraphics[width=0.4\columnwidth]{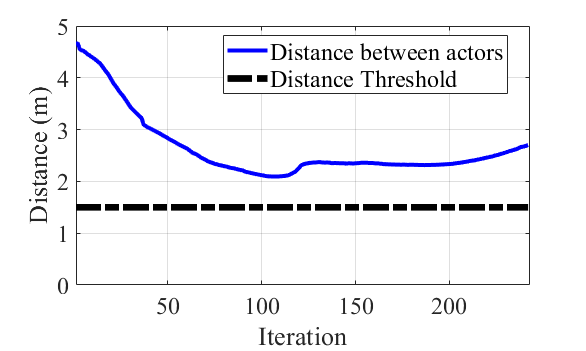}}
	\caption{Snapshots and distances of lab experiment.}
    \label{fig:exp1}
\end{figure}

In the first experiment shown in Fig.~\ref{fig:exp1}, the robot predicts, explains, and takes corrective actions proactively to avoid the human. The robot moves from $(-2.5,0)$m to $(2.5,0)$m at a nominal velocity of $v_r = 0.6$m/s, and the distance threshold $\delta_{th}=1$m.
The robot predicts $\mathcal{P}_h(t) = \lambda$ and explains:\\
\indent$\mathcal{E}_h(t)=$ \textit{\textbf{Interfering}} because: $\{d'<0.21, d_x>2.59\}$\\
With the correction tree, the robot computes:\\
\indent$\bm{\mathcal{C}}_h(t)=$ \textit{\textbf{Not Interfering}} when: $\{v_r \geq 0.5,\ell > 0.5\}$\\
Thus, the corrective action is to maintain nominal velocity, and move to the lane in the positive $y$-direction. 
The minimum distance between actors is $\delta_{\min} = 2.09$m (Fig.~\ref{fig:exp1dist}).

\begin{figure}[ht]
	\centering
	\subfigure[Experiment snapshots before updates. \label{fig:exp2asnaps}]{\includegraphics[width=0.45\linewidth]{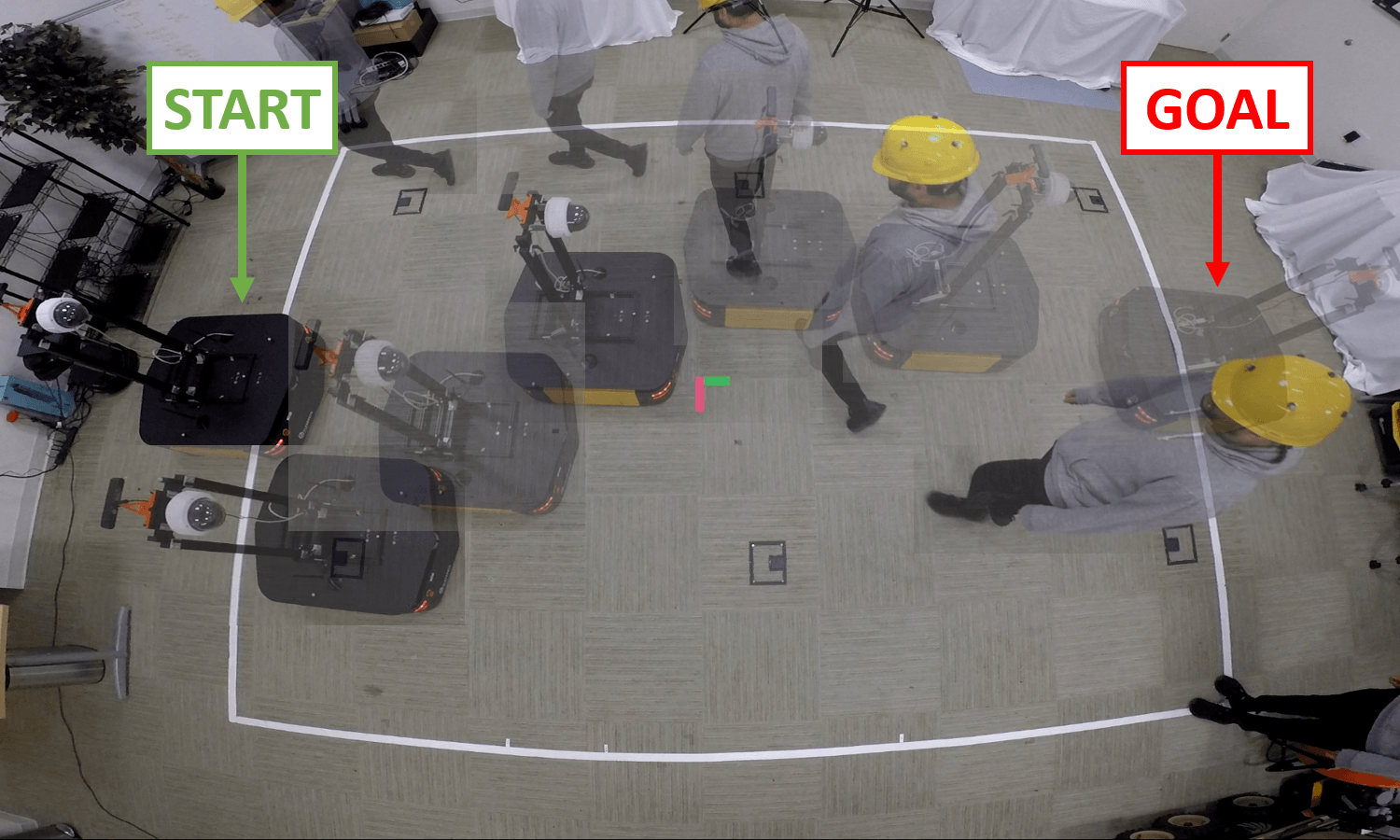}}
	\subfigure[Distance between actors before update. \label{fig:exp2adist}]{\includegraphics[width=0.45\linewidth]{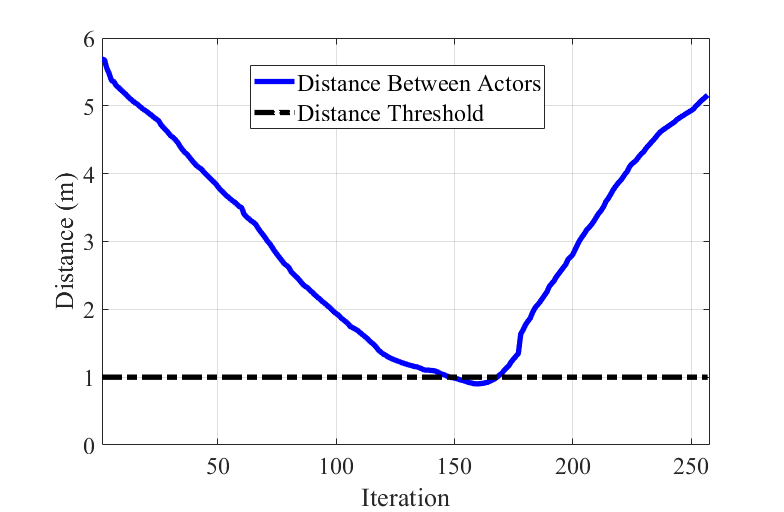}}
	\subfigure[Experiment snapshots after updates. \label{fig:exp2bsnaps}]{\includegraphics[width=0.45\linewidth]{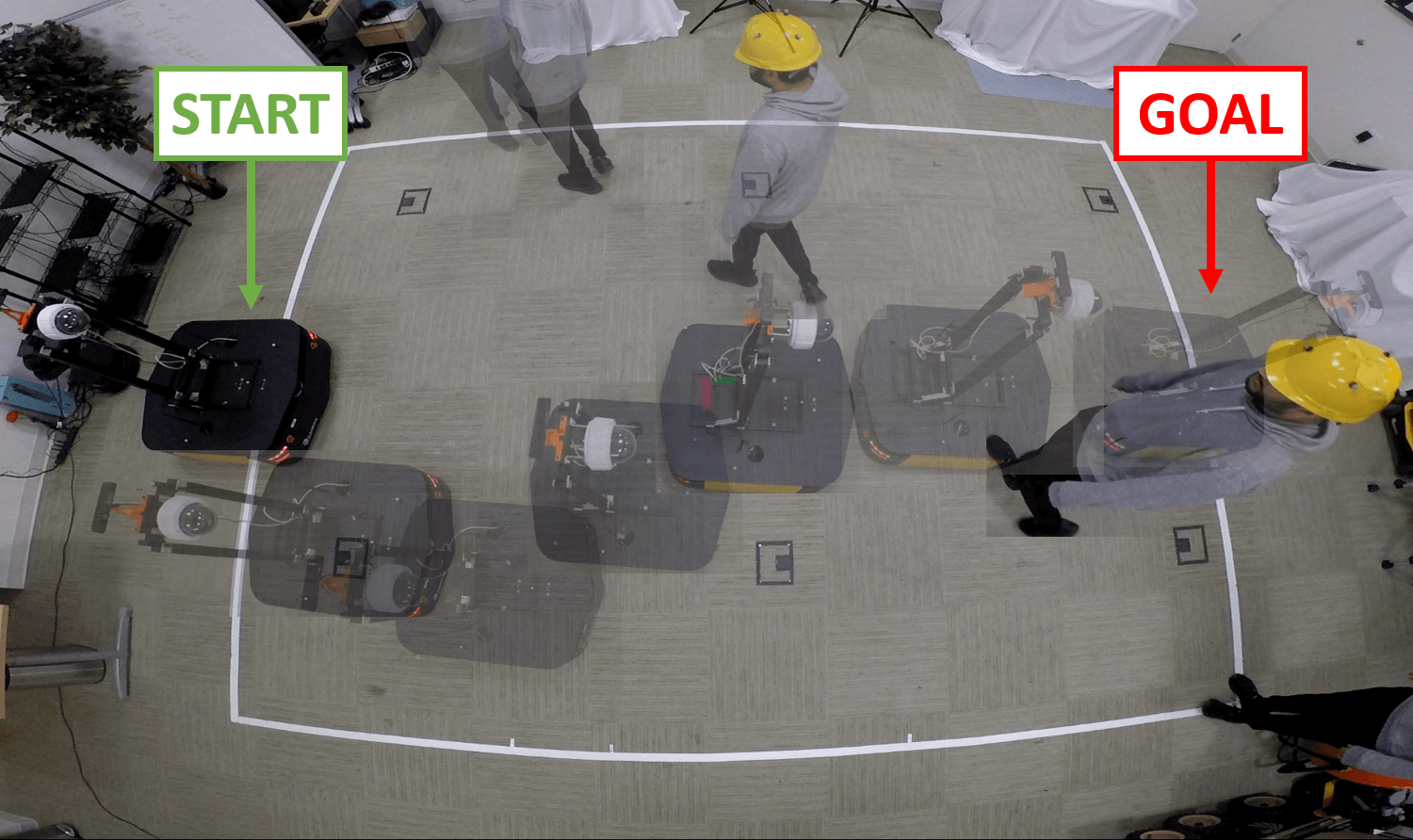}}
	\subfigure[Distance between actors after update. \label{fig:exp2bdist}]{\includegraphics[width=0.45\linewidth]{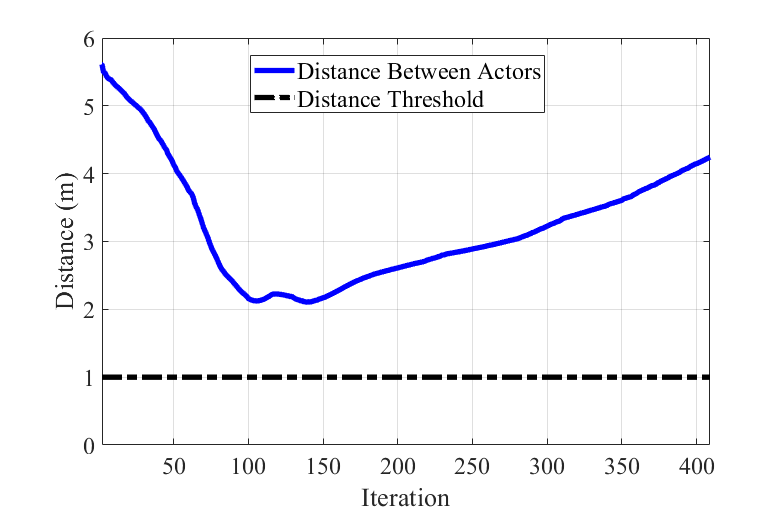}}
	\caption{Experiment showing the effects run-time updates.}
	\label{fig:updateexp}
	\vspace{-10pt}
\end{figure}

In the experiment shown in Fig. \ref{fig:updateexp}, we examine the effects of online validation and updating.
As a proof of concept, we removed a large portion of our training data, and as shown in the first run in Fig.~\ref{fig:exp2asnaps} we observed that the robot moved incorrectly toward the human. The distance threshold is violated, $\delta_{\min} = 0.91$m (Fig.~\ref{fig:exp2adist}), and the person alters his path, reacting to the robot's interfering behavior. After the model is updated and reinforced at run-time, the robot makes the appropriate correction and no interference is observed, with $\delta_{\min} = 2.13$m (Fig.~\ref{fig:updateexp}(c-d)).

\begin{figure}[h]
    \centering
    \subfigure[Initial positions of actors.
	\label{fig:2pplsnap1}]{\includegraphics[width=0.32\columnwidth]{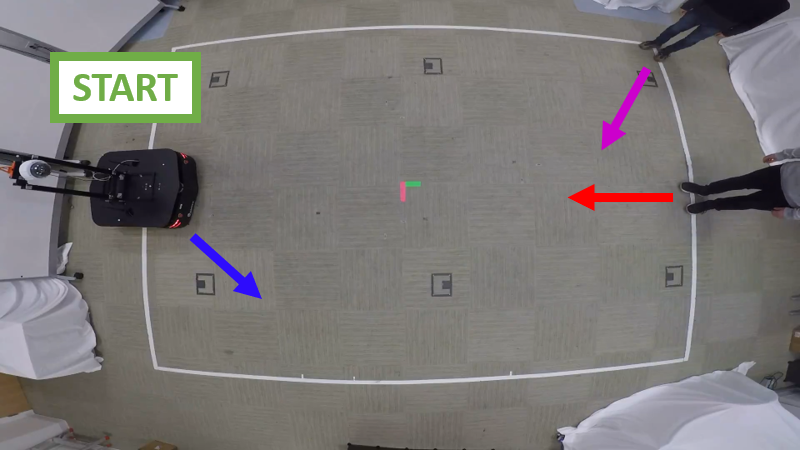}}
    \subfigure[Robot performs corrective behavior.
	\label{fig:2pplsnap2}]{\includegraphics[width=0.32\columnwidth]{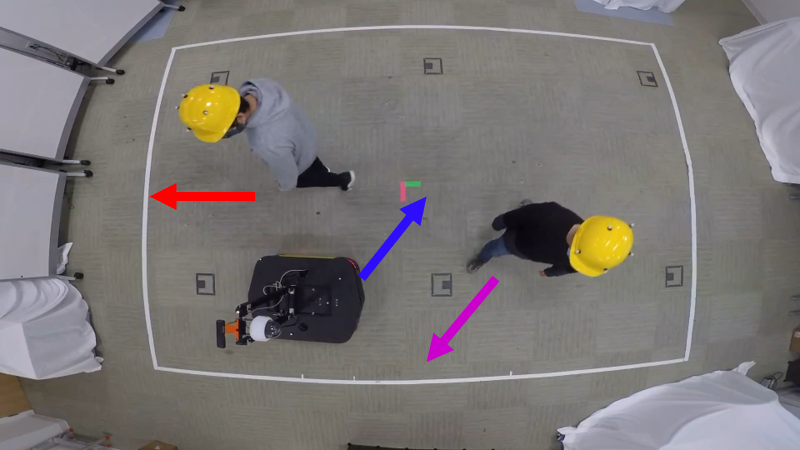}}
    \subfigure[Intermediate positions of actors
	\label{fig:2pplsnap3}]{\includegraphics[width=0.32\columnwidth]{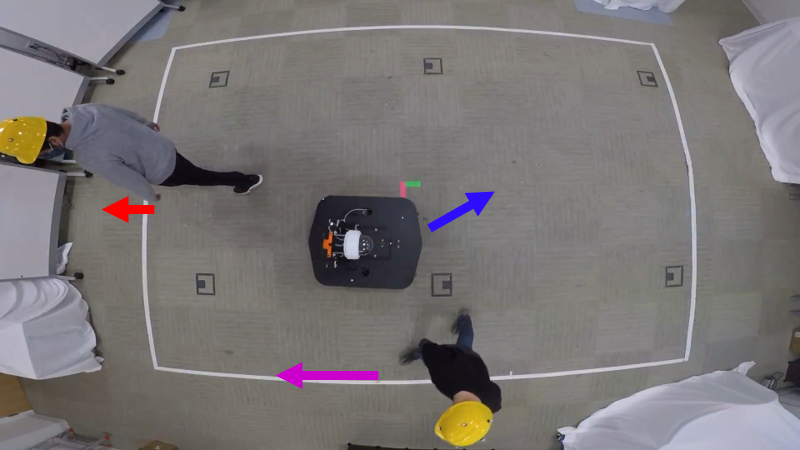}}
	\subfigure[Final positions of actors.
	\label{fig:2pplsnap4}]{\includegraphics[width=0.32\columnwidth]{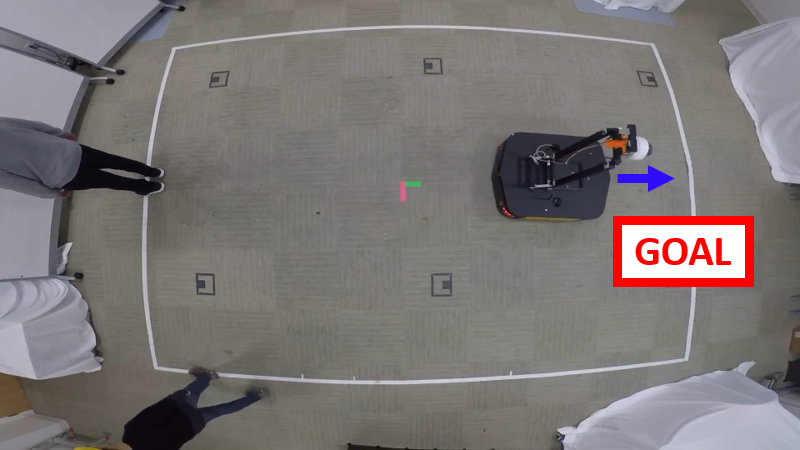}}
	\subfigure[Trajectories of actors.
	\label{fig:2ppltraj}]{\includegraphics[width=0.32\columnwidth]{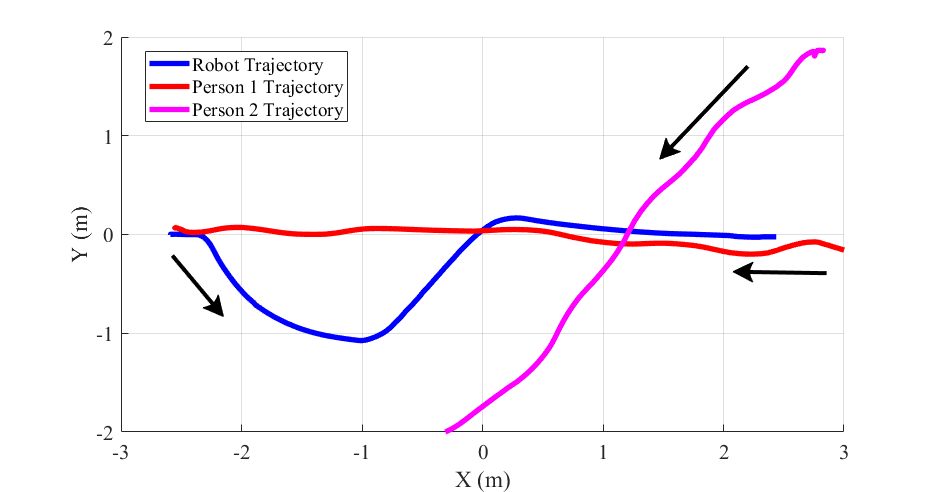}}
	\subfigure[Distance between actors.
	\label{fig:2ppldis}]{\includegraphics[width=0.32\columnwidth]{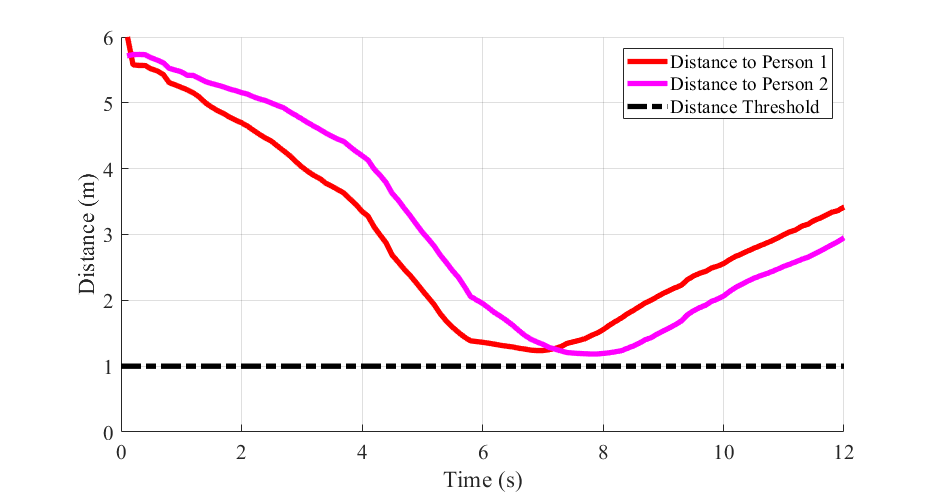}}
	\caption{Snapshots, trajectories and distances of 2-person lab experiment.}
    \label{fig:2ppl}
    \vspace{-10pt}
\end{figure}

In Fig.~\ref{fig:2ppl}, we show the effectiveness of our approach with two people in the lab environment. The robot has the goal to reach $(2.5,0)$m and successfully avoids interference with both people at once, even in a small space, showing that our approach scales to accommodating multiple actors at the same time. The trajectories for all agents are shown in Fig.~\ref{fig:2ppltraj}, and the minimum distance between any human and robot was $1.23$m, which is above the threshold $\delta_{th}=1$m.

\subsubsection{On-Board Sensing Experiment}

\begin{figure}[h]
    \centering
	\subfigure[Initial positions of actors.
	\label{fig:link2pplsnap1}]{\includegraphics[width=0.32\columnwidth]{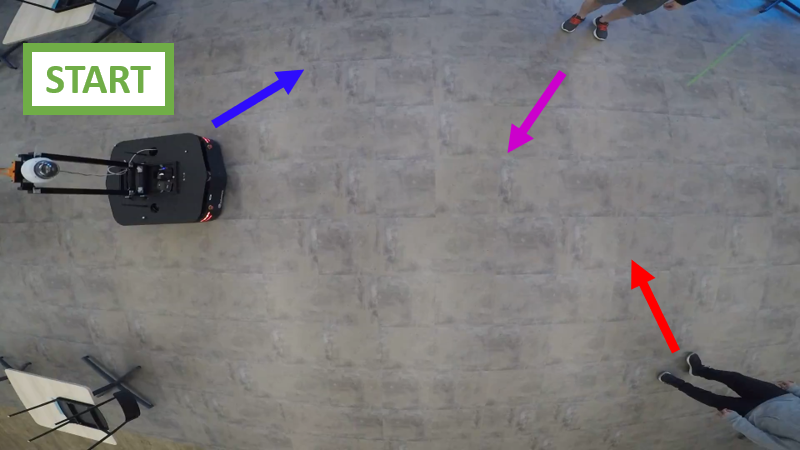}}
    \subfigure[Robot performs corrective behavior.
	\label{fig:link2pplsnap2}]{\includegraphics[width=0.32\columnwidth]{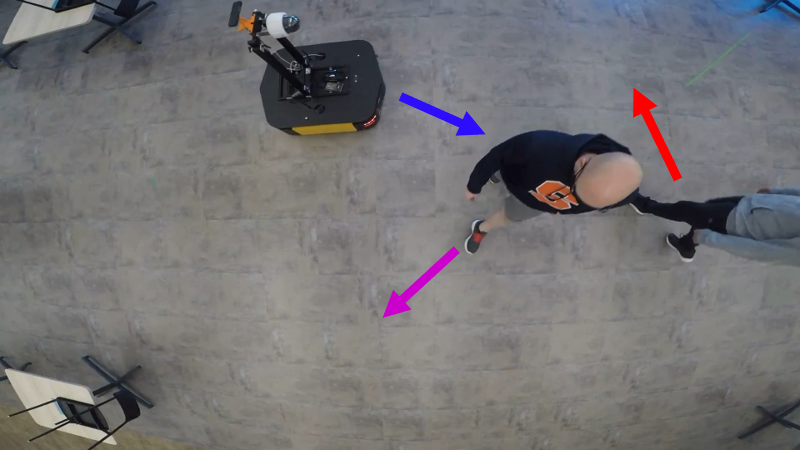}}
    \subfigure[Actors arrive at final positions
	\label{fig:link2pplsnap3}]{\includegraphics[width=0.32\columnwidth]{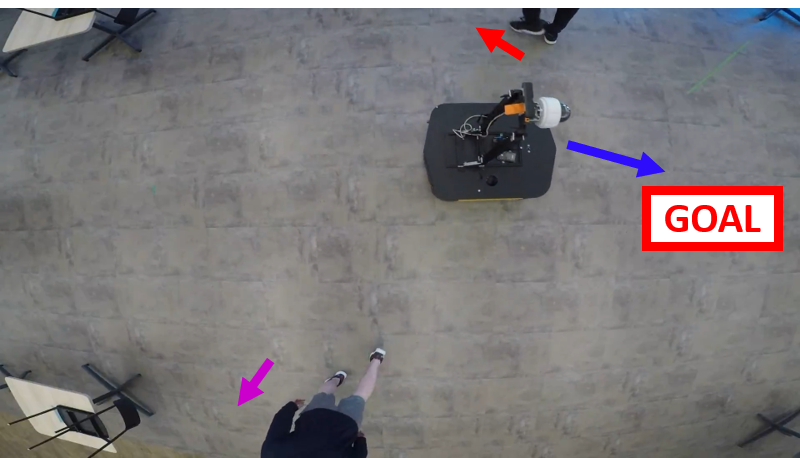}}
	\subfigure[Robot first person view.
	\label{fig:fpv}]{\includegraphics[width=0.32\columnwidth]{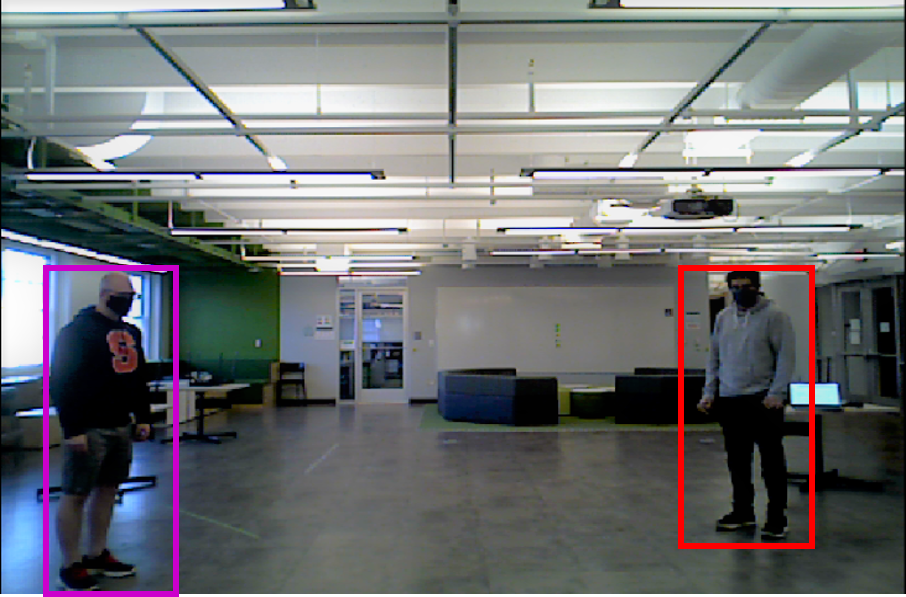}}
	\subfigure[Trajectories of actors.
	\label{fig:linktraj}]{\includegraphics[width=0.32\columnwidth]{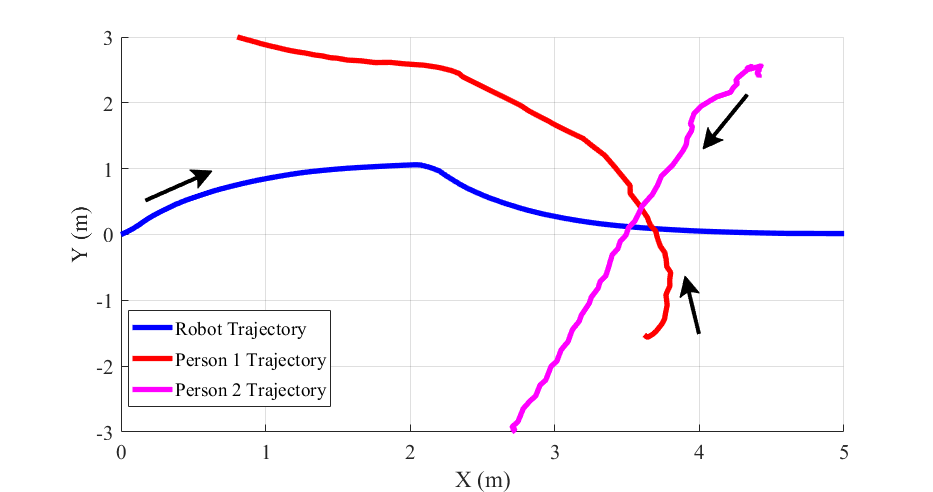}}
	\subfigure[Distance between actors.
	\label{fig:linkdist}]{\includegraphics[width=0.32\columnwidth]{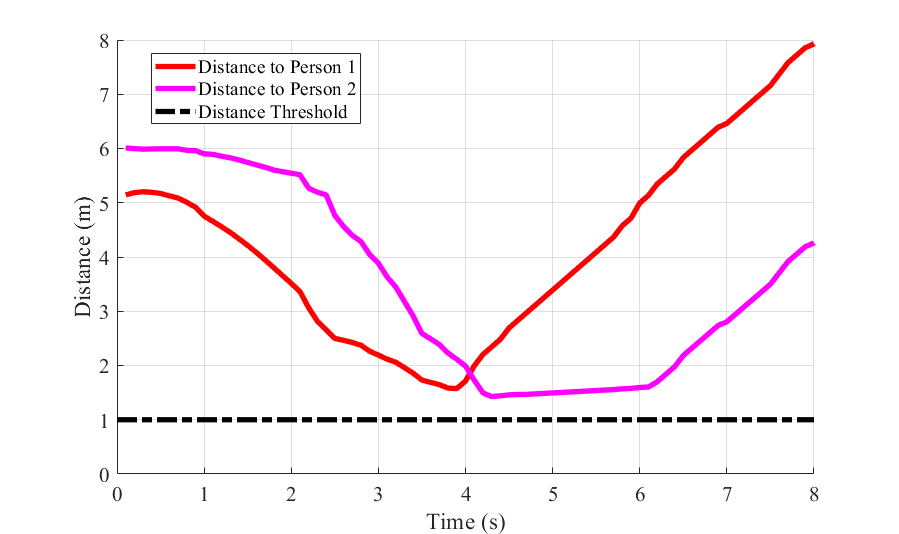}}
	\caption{Snapshots and trajectories of 2-person lab experiment.}
    \label{fig:cam}
    \vspace{-10pt}
\end{figure}

To demonstrate the applicability of our approach outside MOCAP settings, we deployed our technique on the same robot using only the on-board RGB-D and Lidar sensors to identify and track surrounding humans and localize itself. Fig.~\ref{fig:cam} shows the results for this experiment in which the robot is able to successfully avoid interference with surrounding people and reaches its goal without violating the distance threshold despite noisy camera measurements and uncertain person detection and tracking.

\section{Conclusions \& Future Work} \label{sec:concs}

In this work, we have presented a novel approach for interpretable prediction and planning of a robot in a co-robotic environment. We relax the requirement of explicitly predicting human paths, and instead directly predict, explain, and find counterfactual rules for interfering behaviors with binary decision trees and a library of pre-trained trajectories. Unique from other works in this field, we validate robot behaviors to update the predictive model at run-time, resulting in improved behaviors in future operations. While we focused on human-robot operations in this work, our framework works for any path planning operation with multiple actors. The results overall show desirable robot behaviors among humans, though we found that performance can decrease in very dense crowds. 

In current and future work we are exploring the idea of training decision trees using probabilities measured, collected, and updated at run-time to assist decision making in traditional planners and controllers. In addition, we plan to study how learning online can be leveraged safely and efficiently without any reliance on a pre-trained dataset.

\section{Acknowledgement}

This material is based upon work supported by NSF under grant number \#1816591 and 
the Air Force Research Laboratory and the Defense Advanced Research Projects Agency under Contract No. FA8750-18-C-0090.


%


 

\bibliographystyle{IEEEtran}
\bibliography{ms.bib}

\end{document}